\documentclass[letterpaper, 10 pt, conference]{ieeeconf}  

\IEEEoverridecommandlockouts                              

\usepackage{graphics} 
\usepackage{epsfig} 
\usepackage{times} 
\usepackage{amsmath} 
\usepackage{amssymb}  
\usepackage{hyperref}
\usepackage{bm}
\usepackage{cite}
\usepackage{multirow}
\usepackage{wrapfig}
\usepackage[caption=false, font=footnotesize]{subfig}
\usepackage{booktabs}
\usepackage{xcolor}
\usepackage{pifont}
\newcommand{\cmark}{\ding{51}}%
\newcommand{\xmark}{\ding{55}}%
\usepackage{algorithm2e}
\title{\LARGE \bf
CoGrasp: 6-DoF Grasp Generation for Human-Robot Collaboration
}

\author{Abhinav K. Keshari, Hanwen Ren, and Ahmed H. Qureshi 
\thanks{A.K. Keshari, H. Ren, and A.H. Qureshi are with Purdue University,
{\tt\small \{akeshari, ren221, ahqureshi\}@purdue.edu}}
}

\begin{document}

\maketitle
\thispagestyle{empty}
\pagestyle{empty}

\begin{abstract}
Robot grasping is an actively studied area in robotics, mainly focusing on the quality of generated grasps for object manipulation. However, despite advancements, these methods do not consider the human-robot collaboration settings where robots and humans will have to grasp the same objects concurrently. Therefore, generating robot grasps compatible with human preferences of simultaneously holding an object becomes necessary to ensure a safe and natural collaboration experience. In this paper, we propose a novel, deep neural network-based method called CoGrasp that generates human-aware robot grasps by contextualizing human preference models of object grasping into the robot grasp selection process. We validate our approach against existing state-of-the-art robot grasping methods through simulated and real-robot experiments and user studies. In real robot experiments, our method achieves about 88\% success rate in producing stable grasps that also allow humans to interact and grasp objects simultaneously in a socially compliant manner. Furthermore, our user study with 10 independent participants indicated our approach enables a safe, natural, and socially-aware human-robot objects' co-grasping experience compared to a standard robot grasping technique. 
\end{abstract}
\section{INTRODUCTION}
Co-grasping is an essential part of human-robot collaboration tasks where a human and robot simultaneously grasp an object during manipulation. The need for collaborative robot systems has become evident from the lack of available skilled workforce in hospitals, factory floors, and at home to assist people in their daily lives \cite{christensen2021roadmap}. For instance, at hospitals, robots with co-grasping skills can assist in passing various equipment to and from surgeons during surgery or passing medicines to maintain a safe distance between healthcare workers and patients concerning contagious diseases like CoVID-19. Likewise, at factory floors, the tasks for assistive robots could include fetching and handing over various tools to and from their human collaborator in the loop or performing complex assembly tasks through human-machine teaming, which can significantly improve the overall work efficiency and throughput. Similarly,  at home, our elderly often struggle to fetch various objects. Therefore, robots with co-grasping skills can assist them by bringing and handing over different daily-life things, such as utensils, keys, tv remotes, etc. 

\begin{figure}[t]
  \centering
  \subfloat[Human-aware Grasping]{%
\begin{minipage}[b][6cm][t]{.5\columnwidth}
\centering
\includegraphics[width=.95\columnwidth, height=2.9cm]{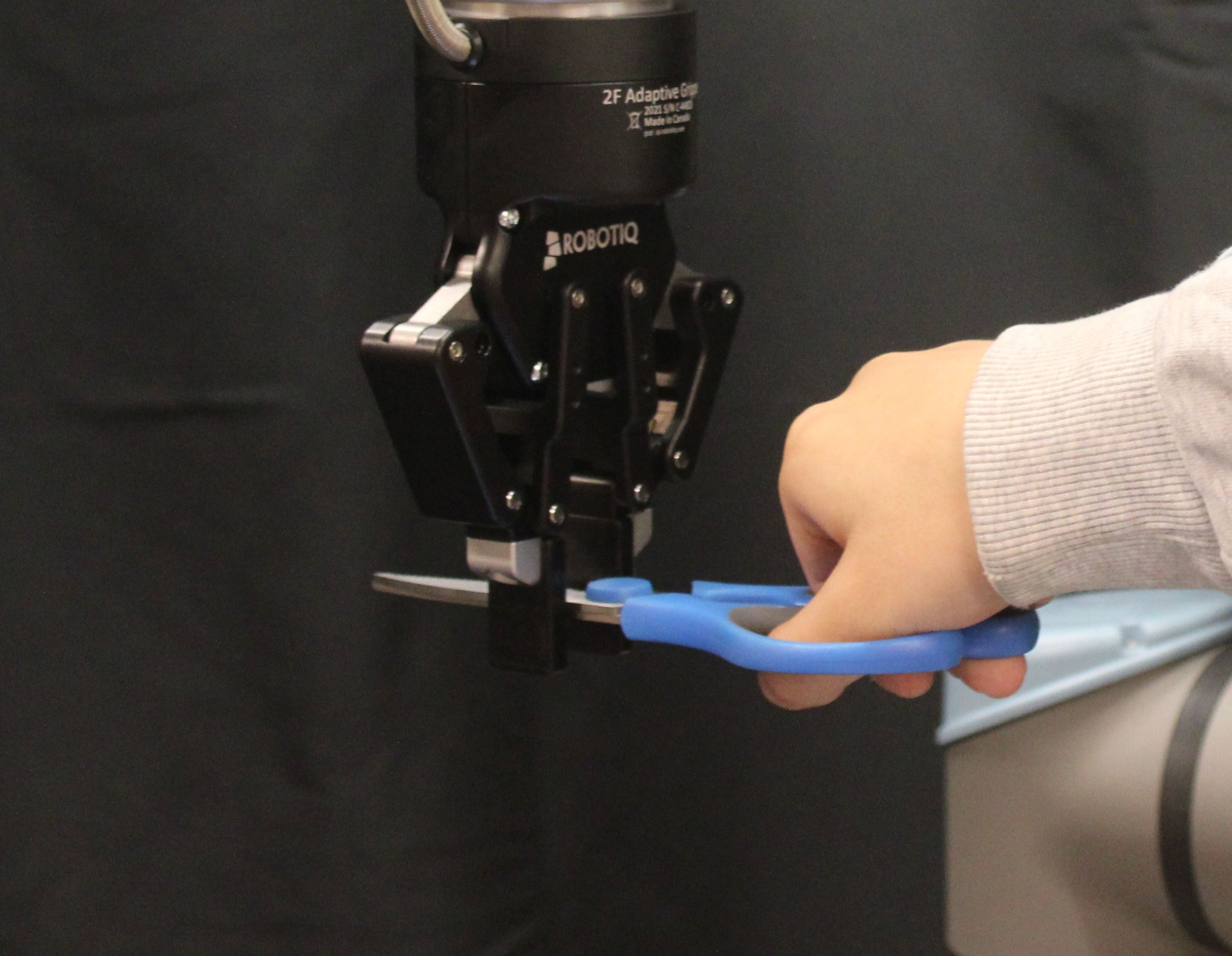}
\vfill
\includegraphics[width=.95\columnwidth, height=2.9cm]{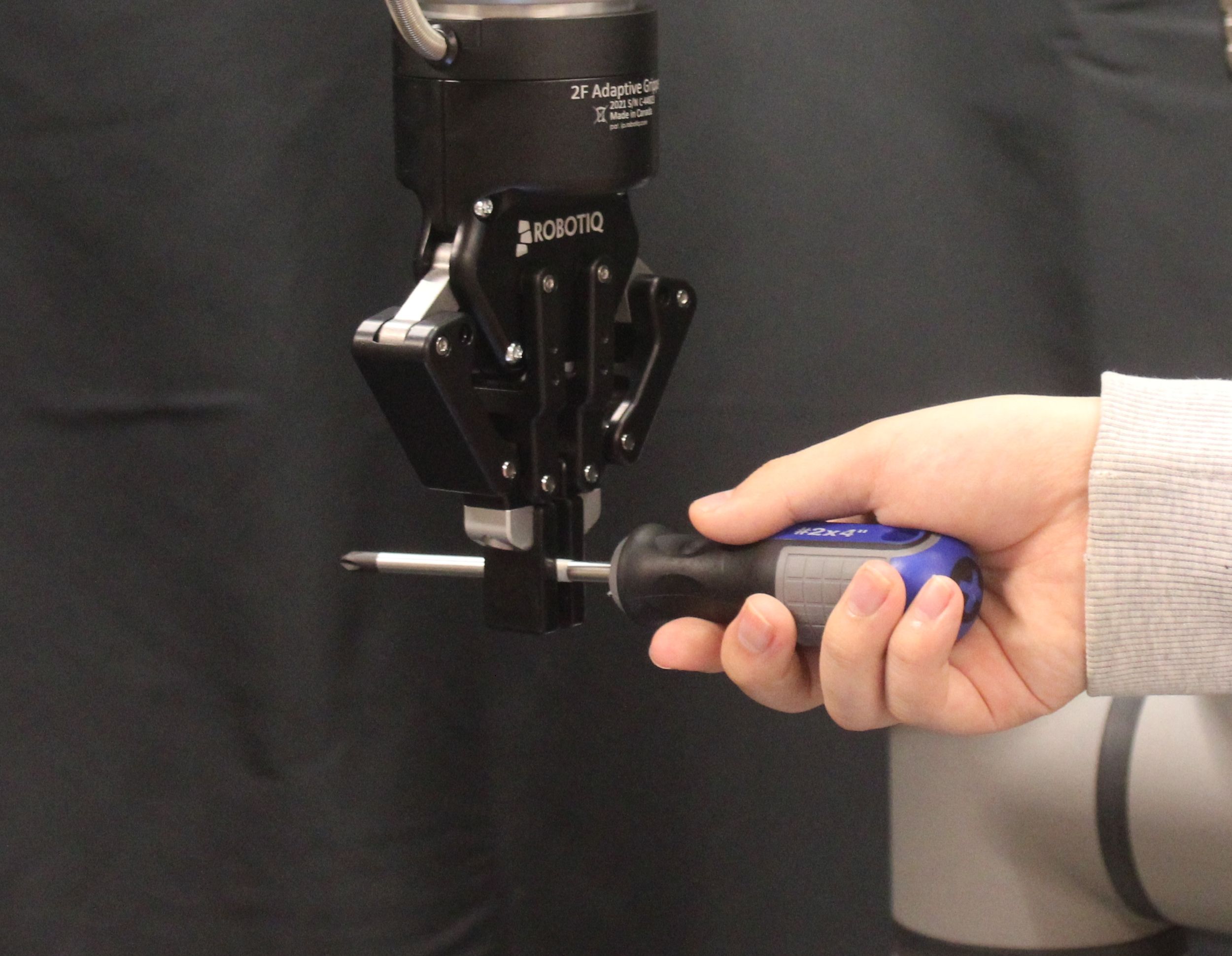}
\end{minipage}
\label{figure1a}}%
  \subfloat[Human-unaware grasping]{%
\begin{minipage}[b][6cm][t]{.5\columnwidth}
\centering
\includegraphics[width=.95\columnwidth, height=2.9cm]{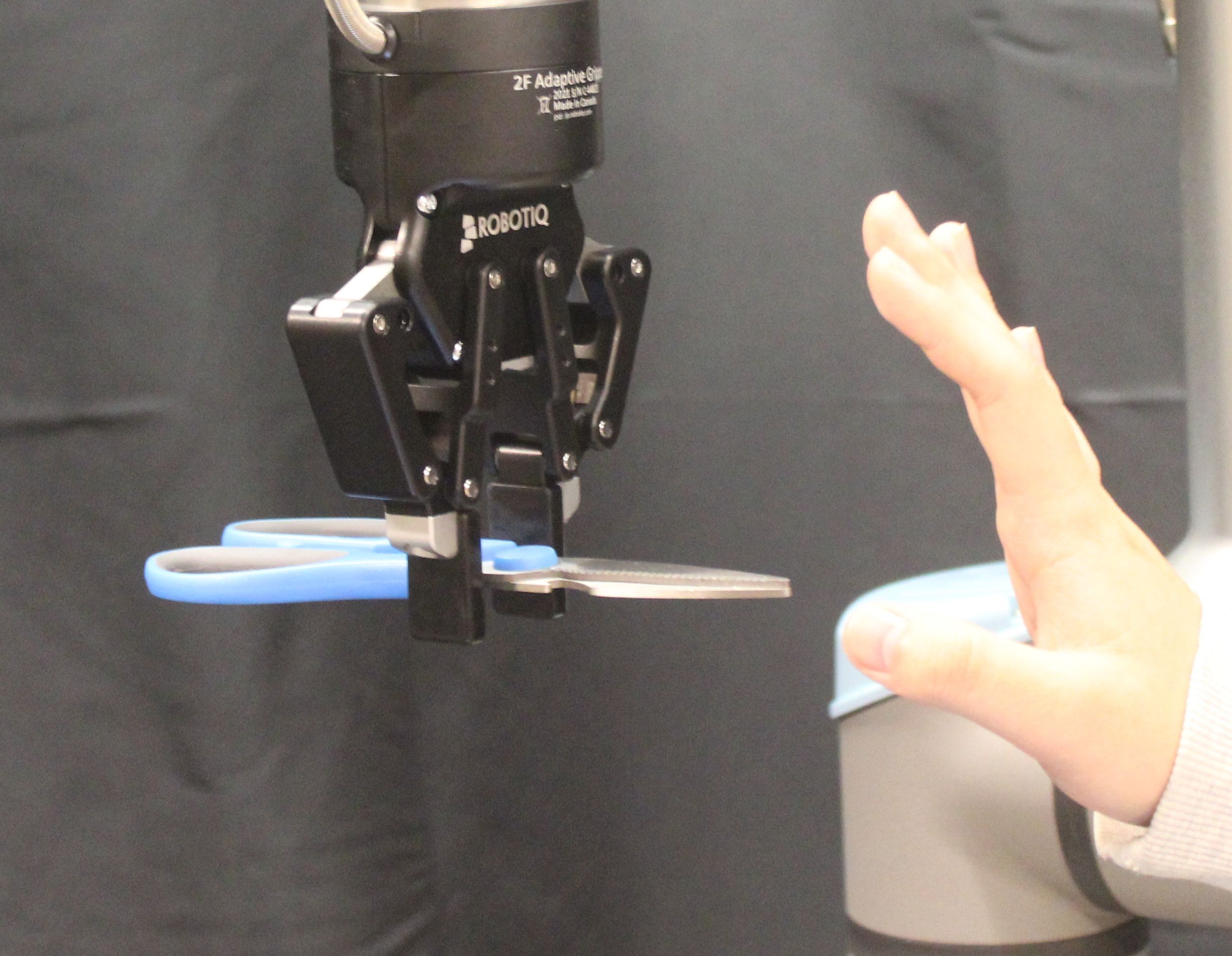}
\vfill
\includegraphics[width=.95\columnwidth, height=2.9cm]{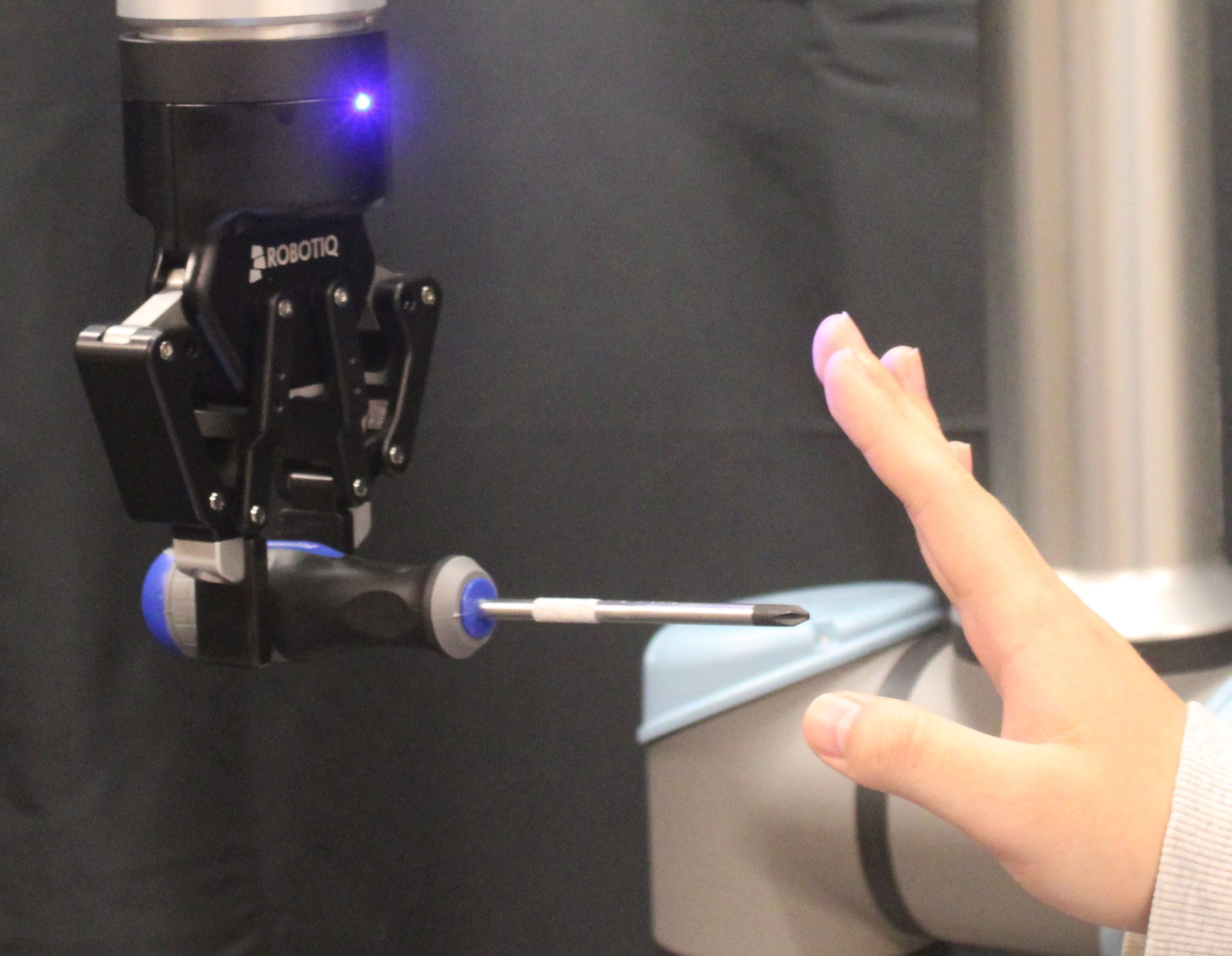}
\end{minipage}
\label{figure1b}}%
  \caption{CoGrasp generates human-aware robot grasps (a) compared to traditional methods (b) that do not consider humans in the loop.}
  \label{figure1}
\vspace{-0.15in}\end{figure}

\begin{figure*}[t]
  \centering
  \includegraphics[trim = {0cm 0cm 0cm 3cm}, width=2.0\columnwidth]{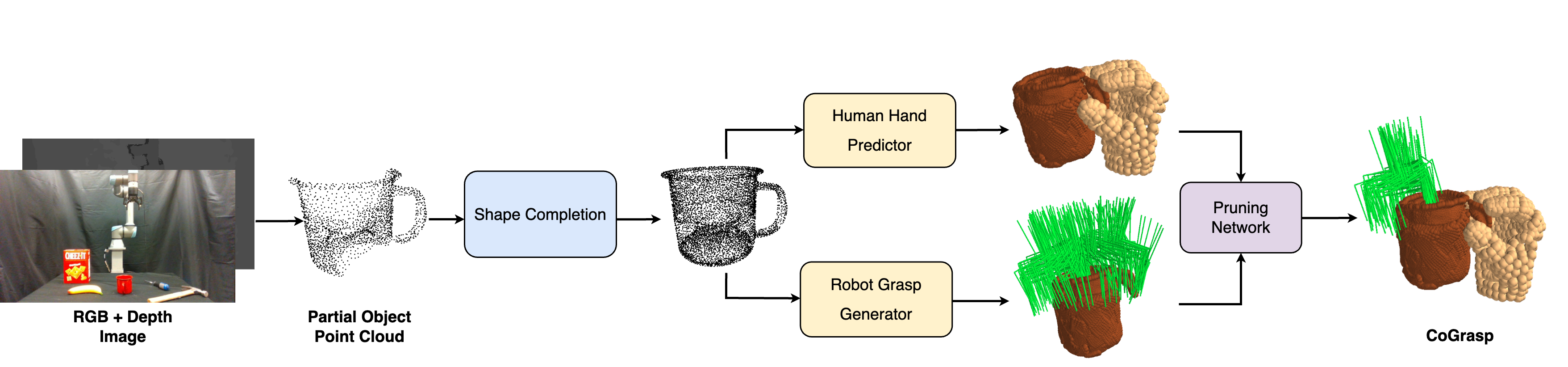}
  \caption{CoGrasp execution pipeline:  Given RGBD information of the scene, our method segments objects partial point clouds and infer their missing parts using the shape completion network. The robot and human hand grasp generators take the completed object point cloud and output the possible robot and human grasp candidates. Finally, the pruning network selects the proper robot grasp compatible for the co-grasping.}
  \label{figure2}
\vspace{-0.15in}\end{figure*}

Although several methods for robot grasp generation exist \cite{ekvall2007learning, liu2020deep, ni2020pointnet++, yan2018learning, sundermeyer2021contact, zeng2018robotic, laskey2016robot, liu2020cage, hoang2022context}, ranging from geometric to data-driven strategies, they are human-agnostic and provide robot-centric algorithms, i.e., observing an object and selecting a gripper’s pose to pick an object without considering collaborating human partners. Generally, in human-robot collaboration, we would expect our robot to grasp things that are also comfortably graspable by their interacting partners during cooperation. For instance, consider a scenario in Fig. \ref{figure1}. Both grasps in Fig. \ref{figure1a} and Fig. \ref{figure1b} would be considered valid for the robot by the existing approaches. However, the grasps in Fig. \ref{figure1b} are inherently invalid as they point sharp ends toward humans, which would be considered unsafe for collaboration. Similarly, in other situations with no sharp objects, the robot would be expected to leave sufficient space for humans to co-grasp objects simultaneously.
In this paper, we propose a human-aware robot grasp generation pipeline called CoGrasp that considers both robot grasp quality and human-in-the-loop for safe and compliant collaboration. Our approach accomplishes human-aware grasping from a raw partial 3D object point cloud by optimizing robot grasp generation using a deep neural network-based object shape completion network, a socially-compliant human grasp prediction network, and a pruning network. Our pruning network builds on our novel co-grasp evaluation algorithm to select stable robot grasps compatible with predicted human grasps for a given object. The overview of our pipeline is shown in Fig. \ref{figure2}. Our approach demonstrates producing grasps appropriate for robot-human collaborative object manipulation, which also works in real-world experiments (Fig. \ref{figure1}). The main contributions of our paper are summarized as follows:

\begin{itemize}
\item A novel and, to the best of our knowledge, the first end-to-end human-aware 6-DoF robot grasp generation method that works in both simulation and real-world environments. 
\item An algorithm that computes grasp quality scores using the geometric information (approach direction and spatial representation) from interactions between the objects, human hand, and robot gripper.
\item A neural model for fast and parallel evaluation of various robot grasp candidates for human friendliness and stability.
\item A new set of metrics that evaluate the quality of grasps based on their safety, human friendliness, and efficiency for human-robot collaboration tasks.
\item A validation of our CoGrasp approach through real-robot experiments, demonstrating our method achieves a 88\% success rate in generating stable robot grasps while leaving socially compliant space on the object for humans to co-grasp concurrently. 
\item A validation of our approach through a user study with 10 participants, indicating our method achieves 22\% higher scores on various metrics of CoGrasp's social compliance and safety than a traditional robot-centric method \cite{sundermeyer2021contact}.
\end{itemize}



\section{Related Work}

This section discusses various techniques that generate collision-free, stable robot grasps for object manipulation. We divide these methods into three categories, i.e., classical, data-driven, and contextual, as described in the following.

\subsection{Classical Method}
The study of robot grasp generation goes very back, starting from attempting to handle objects using a robot hand with elastic fingers\cite{asada1979studies}. It gave rise to geometric-based approaches \cite{nguyen1988constructing, ponce1995computing, ponce1997computing} for producing grasps using the contact points' classification as frictionless, friction, or soft contact to identify parts on the object for a successful grasping. Another line of work, like \cite{schulman2017grasping}, studied the number of contact points needed for stable grasping. The contact points are essential for grasp stability, but the current techniques only look into these for identifying suitable regions from the robot's perspective. In a similar vein, \cite{ferrari1992planning, zheng2012efficient} demonstrated that grasping an object results in a pull force and overcoming that wrench is an essential aspect of stability. \cite{okamura2000overview, bicchi1995closure, prattichizzo1998dynamic} includes studying complex kinematics of the object and the hand motion involved during an interaction, displaying the movement of an approaching hand or the gripper to be critical for grasping. Following the formulations of stable grasping, geometry-based techniques \cite{yan2018learning, bohg2010learning, vahrenkamp2016part, goldfeder2009columbia, li2016dexterous} were proposed that rely directly on the object shape to generate a suitable grasp. However, such methods do not generalize to real-world scenarios where object models are often unknown. Modern methods \cite{liu2020deep} tend to account for surface normals to evaluate the quality of grasps and use them to compute a safe distance for a stable grasp. \cite{zapata2019fast} models the mean axis of an object by running PCA and empirically choosing a safe space from a normal plane to that axis. Despite progress in stable robot grasping, these geometric approaches do not consider human-in-the-loop and solely rely on having one manipulator; thus do not apply to collaboration tasks requiring the co-grasping of the objects.

\subsection{Data-Driven Method}
With the advancement in computational resources and deep neural models, data-driven methods have emerged significantly for generating grasps. In emerging scenarios where input is only available from visual sensors, the existing methods lean on computing the object models and their 6DoF poses \cite{kleeberger2019large} before deploying traditional grasping methods. Furthermore, when the complete 3D object models are unavailable, learning-based shape reconstruction \cite{kleeberger2020single} from inference or multiple views \cite{breyer2021volumetric} are proposed to fill in the gap. Many reinforcement learning techniques have also come up, which learn gripper poses using exploration \cite{quillen2018deep} and learn policies for manipulation \cite{kalashnikov2018scalable}. They help identify dynamic responses to disturbances while grasping. Still, such perturbations are only based on gripper-object and environment interaction. Nevertheless, 3D reconstruction and exploring the large 6-DoF state space suffer from storing a huge amount of data that consists exclusively of the gripper and object relative poses.

Since contact points are crucial in grasping, methods like PointNet++\cite{qi2017pointnet++} are used to learn patterns from point clouds. For instance, \cite{sundermeyer2021contact, liang2019pointnetgpd, ten2017grasp} utilizes PointNet++ to learn geometric forms between grippers and objects from the contact points data available in grasping datasets like ACRONYM \cite{eppner2021acronym}. To extend the point space that is not limited to the contact points, learning-based methods like Dexnet \cite{mahler2017dex, mahler2018dex, fang2018multi} directly learn the orientations and the approach direction of the gripper. There are also methods \cite{fang2020graspnet, qin2020s4g} that tend to produce a grasp score for each point in the space when contact information is not present. However, the contact information and orientations that are looked upon come only from the gripper and the object. Some other neural sampling-based methods \cite{mousavian20196, miller2000graspit} use different grasp quality metrics as objective functions. These metrics depending on the gripper orientations and surface areas, only relate to the overall stability rather than human awareness.

\subsection{Contextual Grasping}
Contextual grasping refers to grasp generation with some context about the objects and their underlying tasks. This problem often involves encoding contextual information like the semantic representation or the object properties into the network inputs. For example, \cite{liu2020cage} encodes the target candidate's visual, tactile, and texture information while performing tasks like picking, lifting, or pouring. \cite{hoang2022context} also considers encoding the relationships of objects in the scene, which allows reasoning about invisible points, enabling collision-free grasp. However, the enhanced reliabilities of grips come from the extensive cost when acquiring hand-labeled training data. In addition, these works focus more on producing human-like grasps or moving the object of interest to the target position, which involves no human actions.

In summary, none of the abovementioned methods considers a simultaneous human grasp while producing a stable robot grasp configuration. The closest idea to finding a suitable grasp for simultaneous grasping may be to learn a policy for ambidextrous grasping \cite{mahler2019learning}. Although this method trains the policy on a large synthetic dataset under the condition where multiple heterogeneous grippers will interact, it does not consider a crucial human-robot interaction task. Furthermore, it can not be directly used for co-grasping as a human will have a preferred way of holding an object and hence the robot grasps used need to be socially compliant for the human to collaborate naturally. 

\section{Proposed Method}
In this section, we present our CoGrasp framework. Given a cluttered scene consisting of unknown objects, we aim to generate robot grasps for all the objects in that scene such that those grasps also allow humans to grasp objects simultaneously in a socially compliant manner. Our method's execution pipeline is shown in Fig. \ref{figure2}, comprising the four main components as described in the following. Furthermore, in this section, we use a notation of $a_{\{B\}}$ to represent any arbitrary set $a$ with $B$ number of elements for brevity.
\subsection{Scene Segmentation \& Shape Completion}
Given a single viewpoint RGBD observation of the scene, we perform image segmentation and extract each object's point cloud by utilizing extrinsic/intrinsic camera parameters. The extracted partial object point clouds via segmentation are denoted as $PC^p_{\{k\}}$ with $k$ instances. These partial point clouds are further processed by our shape completion module, which infers the object's missing surfaces where grasps can be generated. The shape completion helps to maximize the number of potential robot's grasps with a single viewpoint observation. We build this module based on the PoinTr geometry-aware transformer framework \cite{yu2021pointr}. To make the model better generalize to real-world applications, several modifications are introduced. In our setup, we do not assume the object size and its geometric center to be known beforehand. Instead, we use a constant normalization term and the geometric center of the detected partial object point clouds to transfer them into the unit grid before completion. During the completion process, the partial point cloud is first down-sampled and converted to local features using DGCNN \cite{wang2019dynamic} before passing to a transformer-based encoder-decoder module that generates missing points proxies. The completed point cloud for each object is then obtained by giving the missing point proxies to the FoldingNet \cite{yang2018foldingnet}. The rest of our co-grasp framework utilizes the completed points denoted as $PC^c_{\{k\}}$ for robot and human grasp predictions and selections. 
\subsection{Robot Grasp Generator}
This module is used to produce a diverse set of robot grasps $G_r$ for the completed object point clouds $PC^c_{\{k\}}$. In our setup, we leverage a learning-based method called Contact-Graspnet\cite{sundermeyer2021contact} to generate $m\in \mathbb{N}$ number of robot grasp candidates $g_{r\{m\}} \subset G_r$ on each completed object point cloud $PC^c_i$. The Contact-Graspnet's framework consists of PointNet++\cite{qi2017pointnet++} based set abstraction and feature propagation layers, which help the model generalize to real-world sensor data. The output of this module represents each grasp $g_{r}$ using $(R, T) \in SE(3)$ where $R \in SO(3)$ represents the rotation and $T \in \mathbb{R}^{3}$ represents the translation. The training ACRONYM dataset \cite{eppner2021acronym} consists of a diverse set of grasps, enabling it to generalize to out-of-domain objects as well.

\subsection{Human Hand Predictor}
This section presents our human hand grasp prediction module for completed object point clouds $PC^c_{\{k\}}$. In practice, humans have a position preference for holding different objects based on social norms, stability, and safety. For example, we often prefer to hold scissors by the finger rings and mugs by the handles. In our pipeline, this contextual information is provided by the human hand predictor module through learning human hand grasp pose $g_{h}$ on different objects. We use a Variational Autoencoder (VAE) architecture based on \cite{jiang2021hand}. The VAE comprises an encoder-decoder structure. The encoder takes the point clouds of the object $(PC^c_i)$ and the human hand, denoted as $PC^h_{gt}$, as an input and outputs the mean $\boldsymbol{\mu}$ and sigma $\boldsymbol{\sigma}$ in latent space to parametrize the Gaussian Distribution. The decoder takes the latent encoding $z \sim \mathcal{N}(\boldsymbol{\mu}, \boldsymbol{\sigma})$ and the object point cloud $PC^c_i$ as input and outputs the MANO model \cite{MANO:SIGGRAPHASIA:2017} parameters ($\beta$, $\theta$) to represent human hand pose.
The hand parameter $\beta$ estimates the hand shape while the pose parameter $\theta$ describes the rotation and translation of the hand joints. A final MANO differentiable layer gives us the human grasp $g_h$ 6D pose which we render as point cloud $PC^h$. When training the network, we use the ground truth hand point cloud $PC^h_{gt}$ and object point cloud $PC^c_i$ from ObMan dataset \cite{hasson19_obman} as the inputs to produce $\hat{PC^h}$. The objective is to minimize the MSE loss given as $\left \| \hat{PC^h} - PC^h_{gt}\right \|_2 $ and the KL-divergence between $\mathcal{N}(\boldsymbol{\mu}, \boldsymbol{\sigma})$ and $\mathcal{N}(\boldsymbol{0}, \boldsymbol{I})$.
Once trained, 
we use the decoder network that takes a sampled latent code $z\sim (\boldsymbol{0}, \boldsymbol{I})$ and an object point cloud $PC^c_i$ as inputs and outputs the MANO hand parameters. For each object point cloud $PC^c_i$, an $n\in \mathbb{N}$ number of hand grasps $g_{h\{n\}}$ are generated to guide robot grasps selection in the next module.
\subsection{Pruning Network}
Our Pruning Network selects robot grasps $g_{f\{p\}} \subset g_{r\{m\}}$ that are compatible for co-grasping with hand predictions $g_{h\{n\}}$ for each detected object $PC^c_i$. Our network consists of Pointnet++ set abstraction layers which take three point cloud sets $PC^g$, $PC^h$, and $PC^c_i$ representing robot gripper, human hand, and object, respectively, and their feature masks as input and outputs the confidence score $\hat{c} \in [0,1]$. The score $\hat{c}$ indicates the compatibility for a grasp pair ($g_r, g_h$). Recall that $PC^g$ and $PC^h$ are generated by $g_r, g_h$, respectively. Furthermore, we append the feature masks with each point cloud to enable our network to distinguish between them. The feature masks are represented with labels -1, 0, and 1 for the robot gripper, object, and human hand, respectively. 

\begin{figure}%
    \centering
    \subfloat[Large approach angle $\theta_1$]{{\includegraphics[width=4cm]{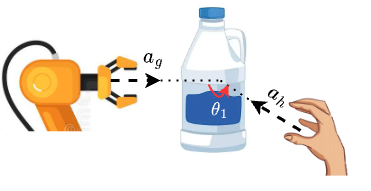} \label{figure3a} }}%
    \subfloat[Small approach angle $\theta_2$]{{\includegraphics[width=3.4cm]{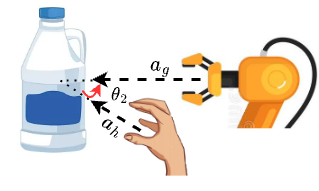} \label{figure3b}}}%
    \caption{A large angle between the approach direction of the robot gripper and human hand makes the co-grasping socially compliant.}%
    \label{figure3}%
\vspace{-0.15in}\end{figure}
\begin{figure*}[t]
  \centering
  \subfloat[Mug]{%
\begin{minipage}[b][6cm][t]{.4\columnwidth}
\centering
\includegraphics[height=3cm]{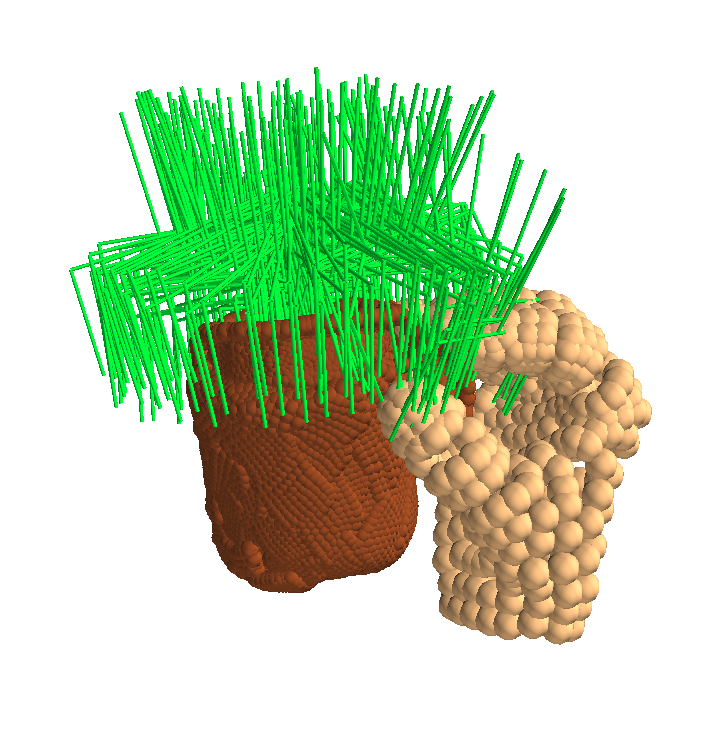}
\vfill
\includegraphics[height=3cm]{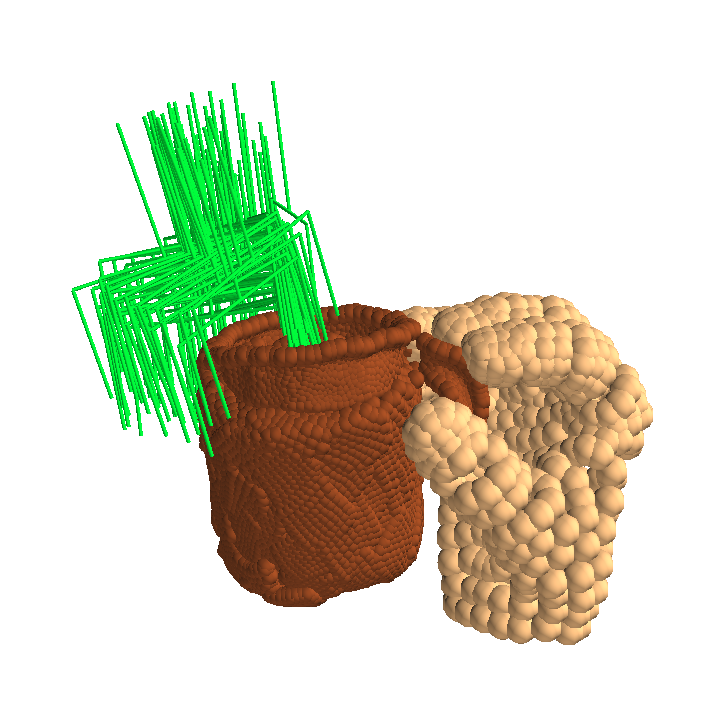}
\end{minipage}}%
  \subfloat[Power Drill]{%
\begin{minipage}[b][6cm][t]{.4\columnwidth}
\centering
\includegraphics[height=3cm]{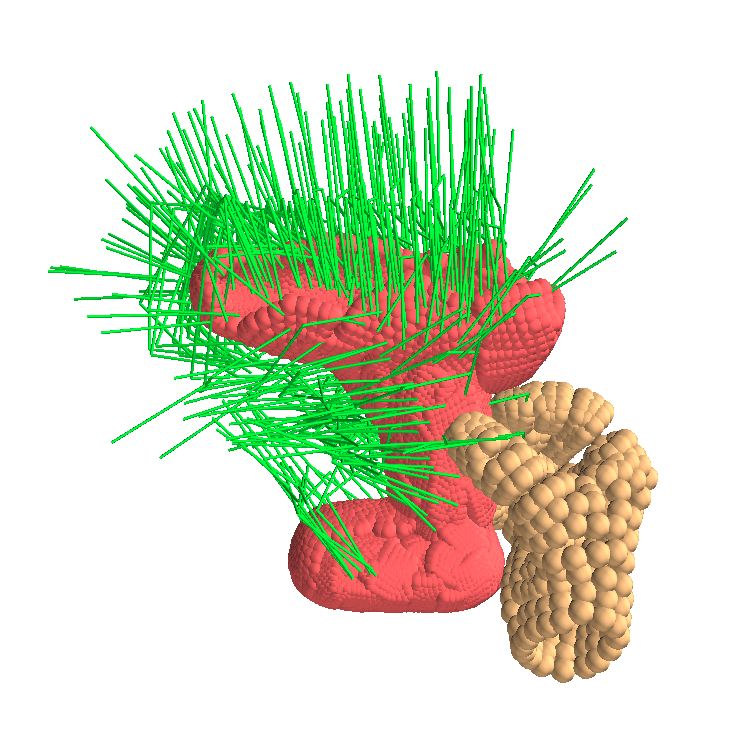}
\vfill
\includegraphics[height=3cm]{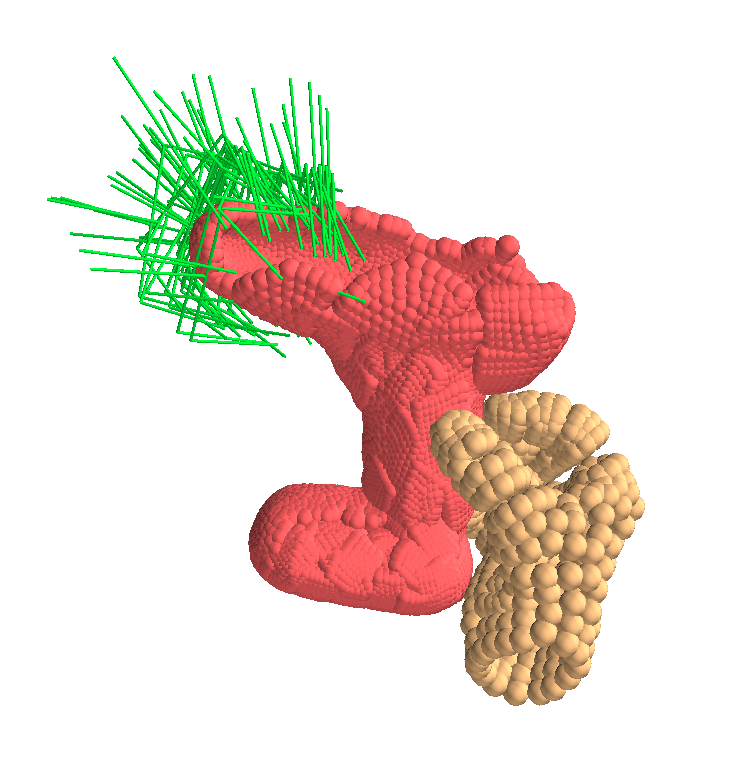}
\end{minipage}}%
  \subfloat[Scissors]{%
\begin{minipage}[b][6cm][t]{.4\columnwidth}
\centering
\includegraphics[height=3cm]{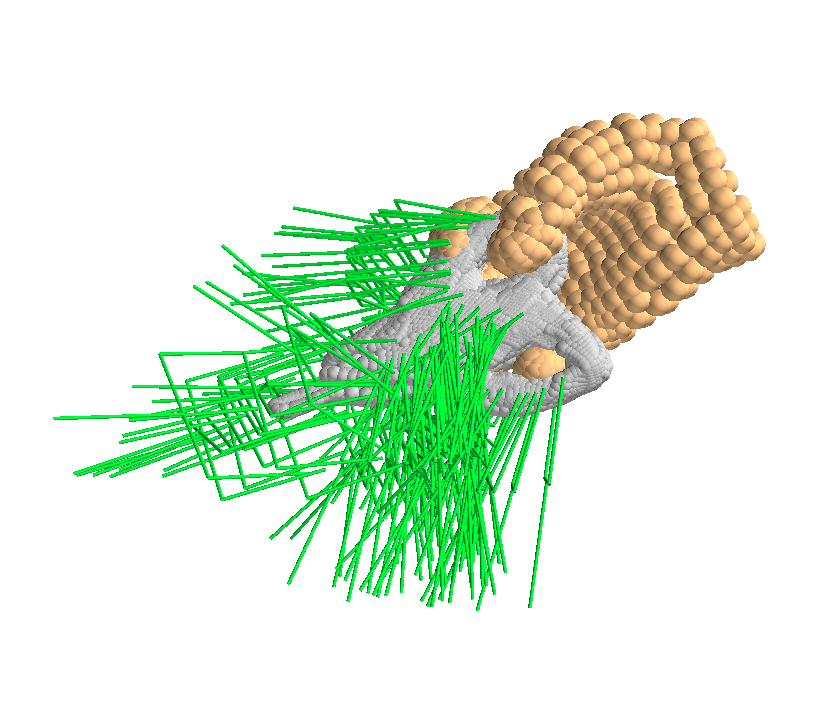}
\vfill
\includegraphics[height=3cm]{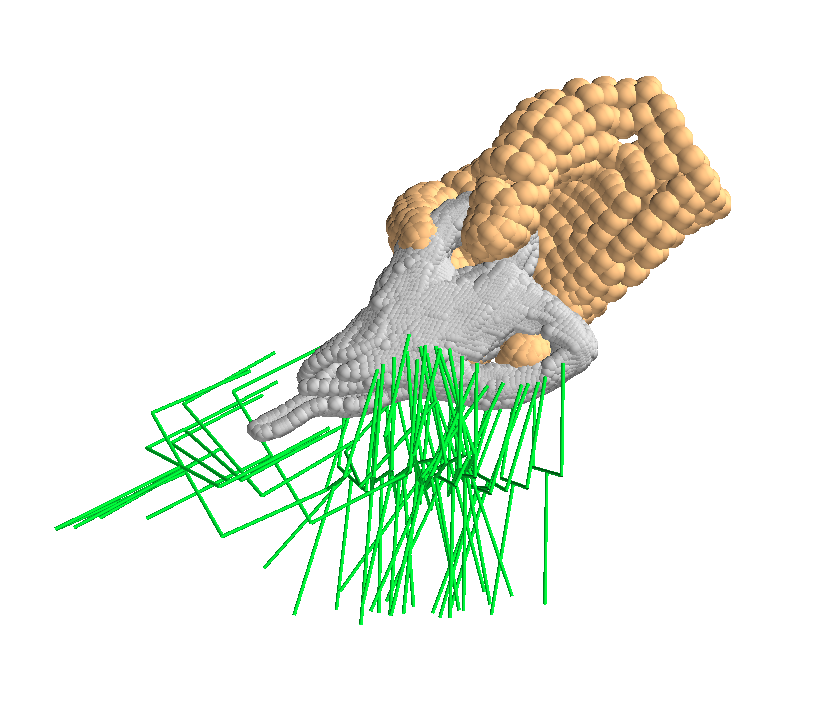}
\end{minipage}}%
  \subfloat[Bleach Cleanser]{%
\begin{minipage}[b][6cm][t]{.4\columnwidth}
\centering
\includegraphics[height=3cm]{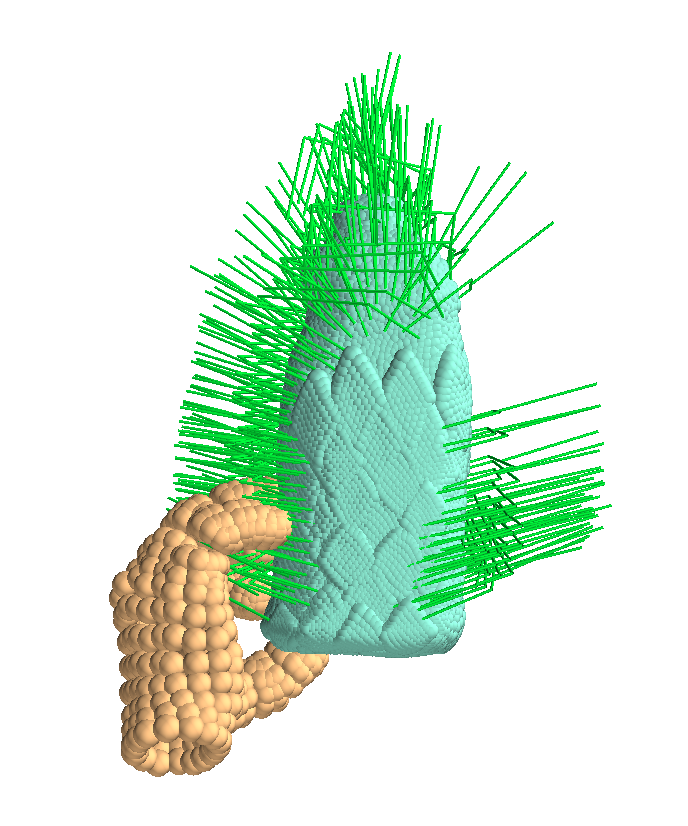}
\vfill
\includegraphics[height=3cm]{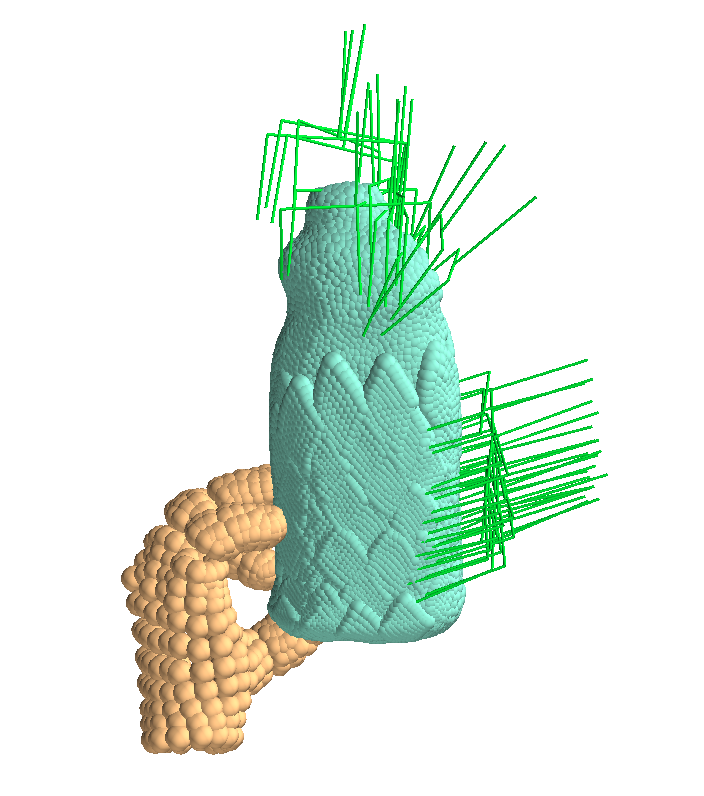}
\end{minipage}}%
  \subfloat[Clamp]{%
\begin{minipage}[b][6cm][t]{.4\columnwidth}
\centering
\includegraphics[height=3cm]{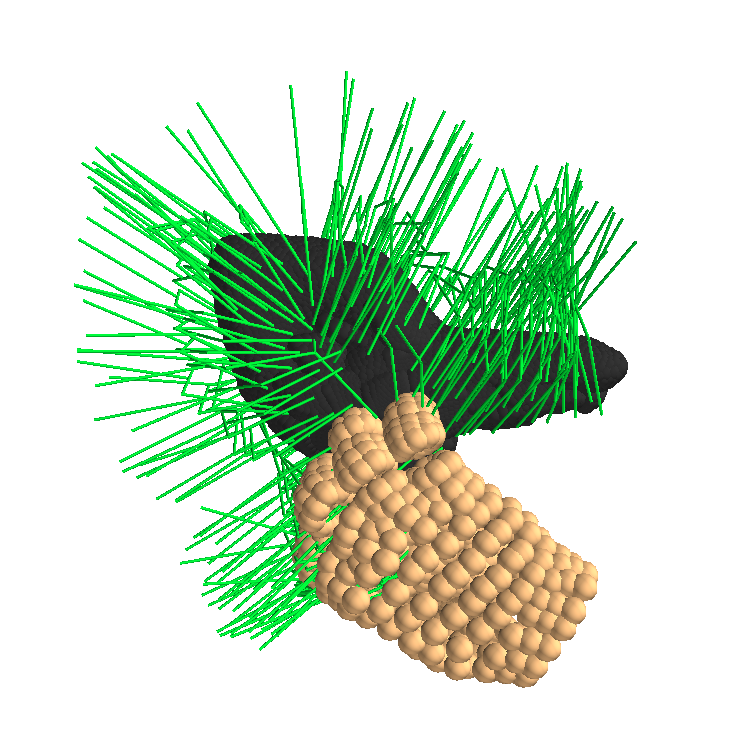}
\vfill
\includegraphics[height=3cm]{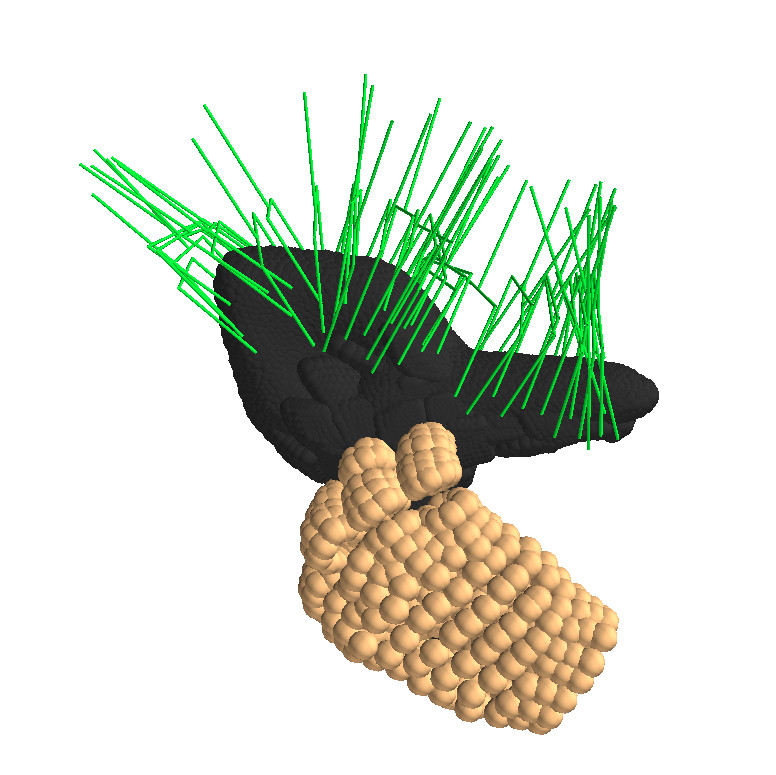}
\end{minipage}}%
  \caption{Simulated images for grasps generated by Contact-GraspNet\cite{sundermeyer2021contact} (top row) and CoGrasp (bottom row) for different objects. We can see that CoGrasp knows how humans will hold an object during collaboration and therefore generates socially compliant robot grasps.}
  \label{figure4}
\end{figure*}

To determine the ground truth confidence scores $c$ for training our pruning network, we define the following two measures:\\ 
\textbf{Distance Measure (}\bm{$S_d$}\textbf{)}: This is used to get an insight into how far apart the gripper and hand are during co-grasping. A more considerable distance ensures the co-grasping can be performed safely. $S_d$ is computed from the average sum of pairwise Euclidean distance between the $PC^g$ and $PC^h$:
$$S_d(PC^g, PC^h) = \frac{\sum _ {\substack{x\in PC^{g} \\ y\in PC^{h}}} \left \| x-y \right \| _{2} }{\left | PC^{g} \right |\left | PC^{h} \right |}  $$\\
\textbf{Angle Measure (}\bm{$S_a$}\textbf{)}:  To increase the co-grasping success rate, the angle between the approach vector of the gripper ($\bm{a_g}$) and the hand ($\bm{a_h}$) needs to be sufficiently large. A large angle assures the robot arm does not collide with the human partner. As illustrated in Fig. \ref{figure3a}, the gripper orientation is complementary to the hand for co-grasping where $\theta_1$ is large. In contrast, Fig. \ref{figure3b} shows that the wrong approach direction makes the robot gripper unnecessarily close to the human. We compute the angle measure $S_a$ as the inner product of the approach vector of the robot and human hand as follows:
$$S_a(\bm{a_g}, \bm{a_h}) = - (\bm{a_g} \cdot \bm{a_h})  $$
To get $\bm{a_{g}}$, we use the representation of $g_{r} \in G_{r}$. We know $g_{r}$ is represented as $(R, T) \in SE(3)$ and its rotation $R$ can be written as follows:
$$R = \begin{bmatrix}
| & | & | \\
\bm{b_g} & \bm{b_g \times a_g}& \bm{a_g} \\
| & | & |\
\end{bmatrix}$$
\\
which gives us $\bm{a_g}$ directly; note $\bm{b_g}$ represents the normalized grasp baseline vector. To get $\bm{a_h}$, we get the set of faces $F_{palm}$ that represents the palm of the MANO\cite{MANO:SIGGRAPHASIA:2017} hand and estimate the approach direction by computing the average of the surface normals from each of these faces. 

To train this network, we generate 89,786 pairs of grasps $(g_r, g_h)$ in simulation (applying 80\%-20\% training/validation split) across all the objects in YCB dataset \cite{calli2017yale}. We consider the grasp pair $(g_r, g_h)$ for an object $i$ as a positive label $c=1$ if its $S_d(PC^g, PC^h) > \lambda_d^i$ and $S_a(\bm{a_g}, \bm{a_h}) > \lambda_a^i$ otherwise negative $c=0$. The thresholds $\lambda_d^i$ and $\lambda_a^i$ are computed as follows. First, we obtain the $m$ robot grasp candidates $g_{r\{m\}}$ and $n$ human grasp candidates $g_{h\{n\}}$ using our robot and human hand grasp generation modules for each object point cloud. This constitutes a set of size $n'=m \times n$ comprising robot and human grasp pairs, i.e., $\{(g_r,g_h)_0, \cdots, (g_r,g_h)_{n'} \}$. Second, for all pairs of the robot and human grasp candidates, we compute the distance $(S_d)$ and angle $(S_a)$ measures resulting in lists $S_D=\{(S_d)_0, \cdots, (S_d)_{n'} \}$ and $S_A=\{(S_a)_0, \cdots, (S_a)_{n'} \}$. Finally, the thresholds $\lambda_d$ and $\lambda_a$ are computed as the median of lists $S_D$ and $S_A$, respectively. 
Given the training dataset with ground truth confidence scores, we train our pruning network using the Binary Cross Entropy (BCE) loss between the predicted confidence score $\hat{c}$ and ground truth $c$. During the evaluation, our pruning network allows us to evaluate predicted grasps in batch, and the final layer being differentiable allows the possibility of refining the grasps by using the gradients $\frac{\partial c}{\partial g_{r}}$.
\section{Experiments}
We run both simulated and real robot experiments to test and compare our method with state-of-the-art grasping methods. 
The simulations are performed in Isaac Gym\cite{makoviychuk2021isaac} physics simulator. For real experiments, we use a UR5e robot manipulator with a 2F-85 Robotiq gripper and Intel RealSense D435i camera for scene observation. 

\subsection{Evaluation Metrics}
Aside from the distance measure $S_d$ and angle measure $S_a$, we also define another metric called the nearest distance measure denoted as $S_n$ to jointly evaluate the level of human awareness of our method's final result. The gripper must not collide with the human hand during the grasping process. This can be checked by measuring the intersection of the convex hull between the gripper and the human hand. Whenever there is an overlap, we set $S_n$ to 0; otherwise, to the nearest distance between the gripper $(PC^{g})$ and human hand $(PC^h)$ point clouds, i.e., $ S_n = \min _ {\substack{x\in PC^{g} \\ y\in PC^{h}}} (\left \| x-y \right \| _{2} ) $. Therefore, a larger $S_n$ reflects the more significant distance between the human and robot grasps for co-grasping.
\subsection{Simulated Experiment}
We simulate CoGrasp, and other state-of-the-art grasping techniques \cite{sundermeyer2021contact, liang2019pointnetgpd} to study large diversified grasping on 92 objects of various categories (Fig. \ref{figure4}). Four different hand grasp poses $g_{h\{4\}}$ are computed for each segmented object $PC^c_i$ per simulation scene to evaluate the proposed gripper poses $g_{f\{p\}}$ from different methods using $S_d$, $S_a$, and $S_n$ metrics introduced above. The results are summarized in Table \ref{table1}, from which we observe better scores for our method in all three measures. Our result is 41\% better than contact-graspnet\cite{sundermeyer2021contact} while 22\% better than PointNetGPD\cite{liang2019pointnetgpd} across all metrics. Overall, the simulation experiments show that our method generates stable, collision-free, and socially compliant grasps needed for human-robot collaboration.

\begin{table}[thpb]
\caption{CoGrasp produces higher quality grasps in terms of metrics $S_a$, $S_d$, and $S_n$ compared to prior grasping methods.}\vspace{-0.15in}
\label{table1}
\begin{center}
\setlength\tabcolsep{4.8pt}
\begin{tabular}{p{2.5cm}ccc}
    \toprule
 & $S_a \uparrow$ &$S_d \uparrow$ & $S_n \uparrow$\\ 
     \midrule
CoGrasp & \textbf{0.675 $\pm$ 0.21} & \textbf{0.129 $\pm$ 0.05} & \textbf{0.037 $\pm$ 0..04} \\ 
Contact-GraspNet\cite{sundermeyer2021contact} & {0.527 $\pm$ 0.24} & {0.117 $\pm$ 0.04} & {0.029 $\pm$ 0.04}  \\ 
PointNetGPD\cite{liang2019pointnetgpd} & {0.492 $\pm$ 0.27} & {0.128 $\pm$ 0.03} & {0.013 $\pm$ 0.02} \\ 
    \bottomrule
\end{tabular} 
\end{center}
\vspace{-0.15in}\end{table}


\subsection{Real Robot Experiment}
In this section, we study the CoGrasp performance in real robot setup in terms of grasp stability and leaving socially-compliant space for humans to co-grasp simultaneously. We create seven different scenes with 19 objects for grasping. The objects were everyday household items, mostly with handles (e.g., screwdriver, scissors, etc.), requiring the robot to leave the handles empty for humans to co-grasp safely. We run three trials for each object and record the results as presented in Table \ref{table2}. To determine the grasp stability, we let the robot move after grasping and observed if the object fell from the gripper or not. To evaluate the human-robot co-grasping suitability, we check whether the standard holding area of the object (e.g., screwdriver handle) is available for humans to grasp concurrently. Our results show that our method exhibits about 88\% success rate over multiple trials in both metrics. The failure cases were mainly related to small objects such as strawberries and clamps where our method could not leave sufficient space for human grasping.
\subsection{Ablation Study}
\begin{figure}[thpb]
  \centering
  \includegraphics[width=0.8\columnwidth]{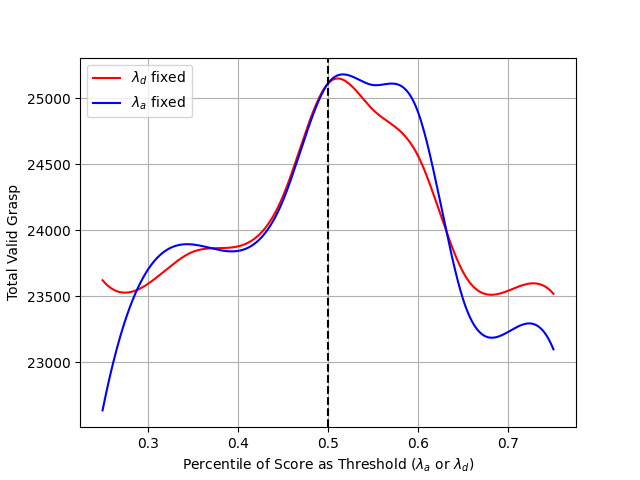}
  \caption{The plot shows the total number of valid grasps generated when keeping $\lambda_{d}$ fixed at the median score and varying $\lambda_{a}$ or vice versa.} 
  \label{figure5}
\vspace{-.15in}\end{figure}

To get diverse human-aware robot grasps, we need to ensure the training data used for Pruning Network is rich in mixed valid grasps. Thus, we want to verify whether the current selection method for positive labels $c$ is reliable. We study the results of our pipeline by selecting different thresholds $\lambda_{d}$ and $\lambda_{a}$ for distance and angle measure, respectively, for computing Pruning Network's labels. We want to assure that the total grasps we can generate for co-grasping are diverse without drastically affecting the quality metrics. To test this, we first fix $\lambda_{a}$ as the median of the angle score and vary $\lambda_{d}$ to compute the total number of grasps that are labeled as positive, and then we do the same with $\lambda_{a}$ by fixing $\lambda_{d}$. The results can be seen in Fig. \ref{figure5}, where we observe that choosing the median score as thresholds for both $\lambda_{d}$ and $\lambda_{a}$ give the most number of positive grasps required for Pruning Networks' robust training.

\begin{table*}[t]
\caption{Real robot experiments results of our method's robot grasps for being stable and socially-compliant for co-grasping.}\vspace{-0.05in}
\label{table2}
\begin{center}
\begin{tabular}{|c|c|c|c|c|c|c|c|c|c|}
    \hline
\multirow{3}{*}{Categories}  & \multirow{3}{*}{Objects} & \multicolumn{4}{c|}{Leaves Human Grabbing Portion} & \multicolumn{4}{c|}{Stability}\\ 
    \cline{3-10} 
  &  & \multicolumn{3}{c|}{Trials} & \multirow{2}{*}{Success Rate} & \multicolumn{3}{c|}{Trials} & \multirow{2}{*}{Success Rate}\\ 
    \cline{3-5} \cline{7-9}
  & & \#1 & \#2 & \#3 & & \#1 & \#2 & \#3 & \\
     \hline
\multirow{3}{*}{Tools} & Hammer & \cmark & \cmark & \cmark & 3/3 & \cmark & \xmark & \cmark & 2/3 \\
& Screw Driver & \cmark & \cmark & \cmark & 3/3 & \cmark & \cmark & \cmark & 3/3 \\
& Large Clamp & \cmark & \cmark & \cmark & 3/3 & \cmark & \cmark  & \cmark  & 3/3 \\
& Medium Clamp & \cmark &  \cmark & \cmark & 3/3 & \cmark & \cmark & \cmark & 3/3 \\
& Small Clamp & \cmark & \xmark & \cmark & 2/3 & \cmark & \cmark  & \xmark & 2/3 \\
& \textbf{Overall} &  &  &  & \textbf{14/15} &  &  &  & \textbf{13/15} \\

    \hline
\multirow{2}{*}{Sharp Objects} & Knife & \cmark & \cmark & \cmark & 3/3 & \xmark & \cmark & \cmark & 2/3 \\
& Scissors & \cmark & \cmark & \cmark & 3/3 & \cmark & \cmark & \cmark & 3/3 \\
& \textbf{Overall} &  &  &  & \textbf{6/6} &  &  &  & \textbf{5/6} \\
\hline
\multirow{2}{*}{Household Objects} & Mug & \cmark & \cmark & \cmark & 3/3 & \xmark & \cmark & \cmark & 2/3 \\
& Fork & \cmark & \cmark & \cmark & 3/3 & \cmark & \cmark & \cmark & 3/3 \\
& Spoon & \cmark & \cmark & \cmark  & 3/3 & \cmark & \cmark & \xmark & 2/3 \\
& Bowl & \cmark & \cmark & \cmark & 3/3 & \cmark & \cmark & \cmark & 3/3 \\
& \textbf{Overall} &  &  &  & \textbf{12/12} &  &  &  & \textbf{10/12} \\
\hline
\multirow{3}{*}{Miscellaneous} & Cracker Box & \cmark & \cmark & \cmark & 3/3 & \cmark & \cmark & \cmark & 3/3 \\
& Banana & \cmark & \cmark & \cmark & 3/3 & \cmark & \cmark & \cmark & 3/3 \\
& Master-Chef Can & \cmark & \cmark & \cmark & 3/3 & \cmark & \cmark & \cmark & 3/3 \\
& Fish Can &  \cmark & \cmark & \cmark & 3/3 &  \cmark& \cmark & \cmark & 3/3 \\
& Orange & \cmark & \cmark & \cmark & 3/3 & \cmark & \cmark & \cmark & 3/3 \\
& Plum & \cmark & \cmark & \cmark & 3/3 & \cmark & \cmark & \cmark & 3/3 \\
& Pear & \cmark & \cmark & \cmark & 3/3 & \cmark & \cmark & \cmark & 3/3 \\
& Strawberry & \xmark & \xmark & \cmark & 1/3 & \cmark & \xmark & \cmark & 2/3 \\
& \textbf{Overall} &  &  &  & \textbf{22/24} &  &  &  & \textbf{23/24} \\
\hline
\end{tabular} 
\end{center}
\vspace{-0.2in}\end{table*}
\subsection{User Study}
We conducted a user study to determine if people could have a more comfortable grasping experience using our method. 10 participants were recruited to interact with the robot for co-grasping using seven household items. The participants were first given a brief time to get accustomed to the testing objects and contemplate how they would like to grasp them. Then the study was divided into two experiments; one executed the grasps predicted by our method, and another involved implementing the grasps generated by Contact-Graspnet\cite{sundermeyer2021contact}. The participants did not know which method the grasps came from, which enforced the unbiased user study. In each study, five objects were arbitrarily selected and used for both experiments. The participants were asked to access the quality of grasps generated from the two methods. After completing the experiments, the participants completed a Likert scale questionnaire describing their overall feedback.

\begin{figure}[thpb]
  \centering
  \includegraphics[width=\columnwidth]{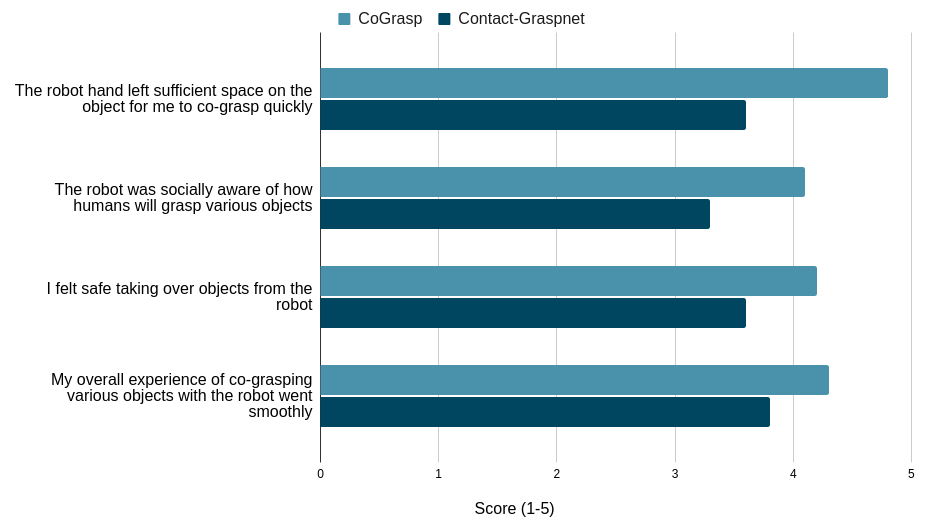}
  \caption{Results from the Likert scale questionnaire given to the users for the study. The average scores for CoGrasp were higher than Contact-Graspnet\cite{sundermeyer2021contact}, demonstrating a socially-compliant co-grasping experience.}
  \label{figure6}
\vspace{-0.15in}\end{figure}

Fig. \ref{figure6} shows the study results of the two experiments summarized from all user responses. For the robot grasps generated by our method, the users could co-grasp all the objects without facing the situation where the robot touches the human hand. On average, the users found the grasps from our method to be socially aware of how human holds the given object and provided similar feedback where one of the users commented \textit{``I can co-grasp the object very easily by the part that human normally uses"}. The users also felt there was enough space to grasp the object in the case of our method compared to \cite{sundermeyer2021contact}. A comment related to that for our method was \textit{``robot left sufficient space for co-grasping"}. 
In addition, our CoGrasp technique was found safer than \cite{sundermeyer2021contact} as one of the users commented during the trial with \cite{sundermeyer2021contact} that they \textit{``felt a bit unsafe grabbing the screw-driver from the sharp side"} signifying the importance of user-friendly grasping.
\section{Conclusion} 
In this paper, we present a novel architecture named CoGrasp that produces the human-aware robot grasps for unknown objects in cluttered environments. Our method generates appropriate robot grasps by checking their compatibility with predicted natural human hand grasps using our novel co-grasping evaluation metrics and pruning function. The results through simulated and real-robot experiments and a user study reflect our design is mindful of handling objects in a socially compliant manner for human-robot collaboration tasks. In our future work, we aim to extend our approach to solve human-robot collaborative object manipulation tasks requiring human-aware robot grasping in the loop. 







\bibliographystyle{IEEEtran}
\bibliography{root}

\begin{thebibliography}{10}
\providecommand{\url}[1]{#1}
\csname url@rmstyle\endcsname
\providecommand{\newblock}{\relax}
\providecommand{\bibinfo}[2]{#2}
\providecommand\BIBentrySTDinterwordspacing{\spaceskip=0pt\relax}
\providecommand\BIBentryALTinterwordstretchfactor{4}
\providecommand\BIBentryALTinterwordspacing{\spaceskip=\fontdimen2\font plus
\BIBentryALTinterwordstretchfactor\fontdimen3\font minus
  \fontdimen4\font\relax}
\providecommand\BIBforeignlanguage[2]{{%
\expandafter\ifx\csname l@#1\endcsname\relax
\typeout{** WARNING: IEEEtran.bst: No hyphenation pattern has been}%
\typeout{** loaded for the language `#1'. Using the pattern for}%
\typeout{** the default language instead.}%
\else
\language=\csname l@#1\endcsname
\fi
#2}}

\bibitem{christensen2021roadmap}
\BIBentryALTinterwordspacing
H.~Christensen, N.~Amato, and H.~Yanco, \emph{A Roadmap for US Robotics - From
  Internet to Robotics 2020 Edition}, ser. Foundations and Trends in Robotics
  Series.\hskip 1em plus 0.5em minus 0.4em\relax Now Publishers, 2021.
  [Online]. Available: \url{https://books.google.com/books?id=zYGPzgEACAAJ}
\BIBentrySTDinterwordspacing

\bibitem{ekvall2007learning}
S.~Ekvall and D.~Kragic, ``Learning and evaluation of the approach vector for
  automatic grasp generation and planning,'' in \emph{Proceedings 2007 IEEE
  International Conference on Robotics and Automation}.\hskip 1em plus 0.5em
  minus 0.4em\relax IEEE, 2007, pp. 4715--4720.

\bibitem{liu2020deep}
M.~Liu, Z.~Pan, K.~Xu, K.~Ganguly, and D.~Manocha, ``Deep differentiable grasp
  planner for high-dof grippers,'' \emph{arXiv preprint arXiv:2002.01530},
  2020.

\bibitem{ni2020pointnet++}
P.~Ni, W.~Zhang, X.~Zhu, and Q.~Cao, ``Pointnet++ grasping: Learning an
  end-to-end spatial grasp generation algorithm from sparse point clouds,'' in
  \emph{2020 IEEE International Conference on Robotics and Automation
  (ICRA)}.\hskip 1em plus 0.5em minus 0.4em\relax IEEE, 2020, pp. 3619--3625.

\bibitem{yan2018learning}
X.~Yan, J.~Hsu, M.~Khansari, Y.~Bai, A.~Pathak, A.~Gupta, J.~Davidson, and
  H.~Lee, ``Learning 6-dof grasping interaction via deep geometry-aware 3d
  representations,'' in \emph{2018 IEEE International Conference on Robotics
  and Automation (ICRA)}.\hskip 1em plus 0.5em minus 0.4em\relax IEEE, 2018,
  pp. 3766--3773.

\bibitem{sundermeyer2021contact}
M.~Sundermeyer, A.~Mousavian, R.~Triebel, and D.~Fox, ``Contact-graspnet:
  Efficient 6-dof grasp generation in cluttered scenes,'' in \emph{2021 IEEE
  International Conference on Robotics and Automation (ICRA)}.\hskip 1em plus
  0.5em minus 0.4em\relax IEEE, 2021, pp. 13\,438--13\,444.

\bibitem{zeng2018robotic}
A.~Zeng, S.~Song, K.-T. Yu, E.~Donlon, F.~R. Hogan, M.~Bauza, D.~Ma, O.~Taylor,
  M.~Liu, E.~Romo, \emph{et~al.}, ``Robotic pick-and-place of novel objects in
  clutter with multi-affordance grasping and cross-domain image matching,'' in
  \emph{2018 IEEE international conference on robotics and automation
  (ICRA)}.\hskip 1em plus 0.5em minus 0.4em\relax IEEE, 2018, pp. 3750--3757.

\bibitem{laskey2016robot}
M.~Laskey, J.~Lee, C.~Chuck, D.~Gealy, W.~Hsieh, F.~T. Pokorny, A.~D. Dragan,
  and K.~Goldberg, ``Robot grasping in clutter: Using a hierarchy of
  supervisors for learning from demonstrations,'' in \emph{2016 IEEE
  international conference on automation science and engineering (CASE)}.\hskip
  1em plus 0.5em minus 0.4em\relax IEEE, 2016, pp. 827--834.

\bibitem{liu2020cage}
W.~Liu, A.~Daruna, and S.~Chernova, ``Cage: Context-aware grasping engine,'' in
  \emph{2020 IEEE International Conference on Robotics and Automation
  (ICRA)}.\hskip 1em plus 0.5em minus 0.4em\relax IEEE, 2020, pp. 2550--2556.

\bibitem{hoang2022context}
D.-C. Hoang, J.~A. Stork, and T.~Stoyanov, ``Context-aware grasp generation in
  cluttered scenes,'' in \emph{2022 International Conference on Robotics and
  Automation (ICRA)}.\hskip 1em plus 0.5em minus 0.4em\relax IEEE, 2022, pp.
  1492--1498.

\bibitem{asada1979studies}
H.~Asada, ``Studies on prehension and handling by robot hands with elastic
  fingers,'' 1979.

\bibitem{nguyen1988constructing}
V.-D. Nguyen, ``Constructing force-closure grasps,'' \emph{The International
  Journal of Robotics Research}, vol.~7, no.~3, pp. 3--16, 1988.

\bibitem{ponce1995computing}
J.~Ponce and B.~Faverjon, ``On computing three-finger force-closure grasps of
  polygonal objects,'' \emph{IEEE Transactions on robotics and automation},
  vol.~11, no.~6, pp. 868--881, 1995.

\bibitem{ponce1997computing}
J.~Ponce, S.~Sullivan, A.~Sudsang, J.-D. Boissonnat, and J.-P. Merlet, ``On
  computing four-finger equilibrium and force-closure grasps of polyhedral
  objects,'' \emph{The International Journal of Robotics Research}, vol.~16,
  no.~1, pp. 11--35, 1997.

\bibitem{schulman2017grasping}
J.~D. Schulman, K.~Goldberg, and P.~Abbeel, ``Grasping and fixturing as
  submodular coverage problems,'' in \emph{Robotics Research}.\hskip 1em plus
  0.5em minus 0.4em\relax Springer, 2017, pp. 571--583.

\bibitem{ferrari1992planning}
C.~Ferrari and J.~F. Canny, ``Planning optimal grasps.'' in \emph{ICRA},
  vol.~3, no.~4, 1992, p.~6.

\bibitem{zheng2012efficient}
Y.~Zheng, ``An efficient algorithm for a grasp quality measure,'' \emph{IEEE
  Transactions on Robotics}, vol.~29, no.~2, pp. 579--585, 2012.

\bibitem{okamura2000overview}
A.~M. Okamura, N.~Smaby, and M.~R. Cutkosky, ``An overview of dexterous
  manipulation,'' in \emph{Proceedings 2000 ICRA. Millennium Conference. IEEE
  International Conference on Robotics and Automation. Symposia Proceedings
  (Cat. No. 00CH37065)}, vol.~1.\hskip 1em plus 0.5em minus 0.4em\relax IEEE,
  2000, pp. 255--262.

\bibitem{bicchi1995closure}
A.~Bicchi, ``On the closure properties of robotic grasping,'' \emph{The
  International Journal of Robotics Research}, vol.~14, no.~4, pp. 319--334,
  1995.

\bibitem{prattichizzo1998dynamic}
D.~Prattichizzo and A.~Bicchi, ``Dynamic analysis of mobility and graspability
  of general manipulation systems,'' \emph{IEEE Transactions on Robotics and
  Automation}, vol.~14, no.~2, pp. 241--258, 1998.

\bibitem{bohg2010learning}
J.~Bohg and D.~Kragic, ``Learning grasping points with shape context,''
  \emph{Robotics and Autonomous Systems}, vol.~58, no.~4, pp. 362--377, 2010.

\bibitem{vahrenkamp2016part}
N.~Vahrenkamp, L.~Westkamp, N.~Yamanobe, E.~E. Aksoy, and T.~Asfour,
  ``Part-based grasp planning for familiar objects,'' in \emph{2016 IEEE-RAS
  16th International Conference on Humanoid Robots (Humanoids)}.\hskip 1em plus
  0.5em minus 0.4em\relax IEEE, 2016, pp. 919--925.

\bibitem{goldfeder2009columbia}
C.~Goldfeder, M.~Ciocarlie, H.~Dang, and P.~K. Allen, ``The columbia grasp
  database,'' in \emph{2009 IEEE international conference on robotics and
  automation}.\hskip 1em plus 0.5em minus 0.4em\relax IEEE, 2009, pp.
  1710--1716.

\bibitem{li2016dexterous}
M.~Li, K.~Hang, D.~Kragic, and A.~Billard, ``Dexterous grasping under shape
  uncertainty,'' \emph{Robotics and Autonomous Systems}, vol.~75, pp. 352--364,
  2016.

\bibitem{zapata2019fast}
B.~S. Zapata-Impata, P.~Gil, J.~Pomares, and F.~Torres, ``Fast geometry-based
  computation of grasping points on three-dimensional point clouds,''
  \emph{International Journal of Advanced Robotic Systems}, vol.~16, no.~1, p.
  1729881419831846, 2019.

\bibitem{kleeberger2019large}
K.~Kleeberger, C.~Landgraf, and M.~F. Huber, ``Large-scale 6d object pose
  estimation dataset for industrial bin-picking,'' in \emph{2019 IEEE/RSJ
  International Conference on Intelligent Robots and Systems (IROS)}.\hskip 1em
  plus 0.5em minus 0.4em\relax IEEE, 2019, pp. 2573--2578.

\bibitem{kleeberger2020single}
K.~Kleeberger and M.~F. Huber, ``Single shot 6d object pose estimation,'' in
  \emph{2020 IEEE International Conference on Robotics and Automation
  (ICRA)}.\hskip 1em plus 0.5em minus 0.4em\relax IEEE, 2020, pp. 6239--6245.

\bibitem{breyer2021volumetric}
M.~Breyer, J.~J. Chung, L.~Ott, R.~Siegwart, and J.~Nieto, ``Volumetric
  grasping network: Real-time 6 dof grasp detection in clutter,'' \emph{arXiv
  preprint arXiv:2101.01132}, 2021.

\bibitem{quillen2018deep}
D.~Quillen, E.~Jang, O.~Nachum, C.~Finn, J.~Ibarz, and S.~Levine, ``Deep
  reinforcement learning for vision-based robotic grasping: A simulated
  comparative evaluation of off-policy methods,'' in \emph{2018 IEEE
  International Conference on Robotics and Automation (ICRA)}.\hskip 1em plus
  0.5em minus 0.4em\relax IEEE, 2018, pp. 6284--6291.

\bibitem{kalashnikov2018scalable}
D.~Kalashnikov, A.~Irpan, P.~Pastor, J.~Ibarz, A.~Herzog, E.~Jang, D.~Quillen,
  E.~Holly, M.~Kalakrishnan, V.~Vanhoucke, \emph{et~al.}, ``Scalable deep
  reinforcement learning for vision-based robotic manipulation,'' in
  \emph{Conference on Robot Learning}.\hskip 1em plus 0.5em minus 0.4em\relax
  PMLR, 2018, pp. 651--673.

\bibitem{qi2017pointnet++}
C.~R. Qi, L.~Yi, H.~Su, and L.~J. Guibas, ``Pointnet++: Deep hierarchical
  feature learning on point sets in a metric space,'' \emph{Advances in neural
  information processing systems}, vol.~30, 2017.

\bibitem{liang2019pointnetgpd}
H.~Liang, X.~Ma, S.~Li, M.~G{\"o}rner, S.~Tang, B.~Fang, F.~Sun, and J.~Zhang,
  ``Pointnetgpd: Detecting grasp configurations from point sets,'' in
  \emph{2019 International Conference on Robotics and Automation (ICRA)}.\hskip
  1em plus 0.5em minus 0.4em\relax IEEE, 2019, pp. 3629--3635.

\bibitem{ten2017grasp}
A.~ten Pas, M.~Gualtieri, K.~Saenko, and R.~Platt, ``Grasp pose detection in
  point clouds,'' \emph{The International Journal of Robotics Research},
  vol.~36, no. 13-14, pp. 1455--1473, 2017.

\bibitem{eppner2021acronym}
C.~Eppner, A.~Mousavian, and D.~Fox, ``Acronym: A large-scale grasp dataset
  based on simulation,'' in \emph{2021 IEEE International Conference on
  Robotics and Automation (ICRA)}.\hskip 1em plus 0.5em minus 0.4em\relax IEEE,
  2021, pp. 6222--6227.

\bibitem{mahler2017dex}
J.~Mahler, J.~Liang, S.~Niyaz, M.~Laskey, R.~Doan, X.~Liu, J.~A. Ojea, and
  K.~Goldberg, ``Dex-net 2.0: Deep learning to plan robust grasps with
  synthetic point clouds and analytic grasp metrics,'' \emph{arXiv preprint
  arXiv:1703.09312}, 2017.

\bibitem{mahler2018dex}
J.~Mahler, M.~Matl, X.~Liu, A.~Li, D.~Gealy, and K.~Goldberg, ``Dex-net 3.0:
  Computing robust vacuum suction grasp targets in point clouds using a new
  analytic model and deep learning,'' in \emph{2018 IEEE International
  Conference on robotics and automation (ICRA)}.\hskip 1em plus 0.5em minus
  0.4em\relax IEEE, 2018, pp. 5620--5627.

\bibitem{fang2018multi}
K.~Fang, Y.~Bai, S.~Hinterstoisser, S.~Savarese, and M.~Kalakrishnan,
  ``Multi-task domain adaptation for deep learning of instance grasping from
  simulation,'' in \emph{2018 IEEE International Conference on Robotics and
  Automation (ICRA)}.\hskip 1em plus 0.5em minus 0.4em\relax IEEE, 2018, pp.
  3516--3523.

\bibitem{fang2020graspnet}
H.-S. Fang, C.~Wang, M.~Gou, and C.~Lu, ``Graspnet-1billion: A large-scale
  benchmark for general object grasping,'' in \emph{Proceedings of the IEEE/CVF
  conference on computer vision and pattern recognition}, 2020, pp.
  11\,444--11\,453.

\bibitem{qin2020s4g}
Y.~Qin, R.~Chen, H.~Zhu, M.~Song, J.~Xu, and H.~Su, ``S4g: Amodal single-view
  single-shot se (3) grasp detection in cluttered scenes,'' in \emph{Conference
  on robot learning}.\hskip 1em plus 0.5em minus 0.4em\relax PMLR, 2020, pp.
  53--65.

\bibitem{mousavian20196}
A.~Mousavian, C.~Eppner, and D.~Fox, ``6-dof graspnet: Variational grasp
  generation for object manipulation,'' in \emph{Proceedings of the IEEE/CVF
  International Conference on Computer Vision}, 2019, pp. 2901--2910.

\bibitem{miller2000graspit}
A.~T. Miller and P.~K. Allen, ``Graspit!: A versatile simulator for grasp
  analysis,'' in \emph{ASME International Mechanical Engineering Congress and
  Exposition}, vol. 26652.\hskip 1em plus 0.5em minus 0.4em\relax American
  Society of Mechanical Engineers, 2000, pp. 1251--1258.

\bibitem{mahler2019learning}
J.~Mahler, M.~Matl, V.~Satish, M.~Danielczuk, B.~DeRose, S.~McKinley, and
  K.~Goldberg, ``Learning ambidextrous robot grasping policies,'' \emph{Science
  Robotics}, vol.~4, no.~26, p. eaau4984, 2019.

\bibitem{yu2021pointr}
X.~Yu, Y.~Rao, Z.~Wang, Z.~Liu, J.~Lu, and J.~Zhou, ``Pointr: Diverse point
  cloud completion with geometry-aware transformers,'' in \emph{Proceedings of
  the IEEE/CVF international conference on computer vision}, 2021, pp.
  12\,498--12\,507.

\bibitem{wang2019dynamic}
Y.~Wang, Y.~Sun, Z.~Liu, S.~E. Sarma, M.~M. Bronstein, and J.~M. Solomon,
  ``Dynamic graph cnn for learning on point clouds,'' \emph{Acm Transactions On
  Graphics (tog)}, vol.~38, no.~5, pp. 1--12, 2019.

\bibitem{yang2018foldingnet}
Y.~Yang, C.~Feng, Y.~Shen, and D.~Tian, ``Foldingnet: Point cloud auto-encoder
  via deep grid deformation,'' in \emph{Proceedings of the IEEE conference on
  computer vision and pattern recognition}, 2018, pp. 206--215.

\bibitem{jiang2021hand}
H.~Jiang, S.~Liu, J.~Wang, and X.~Wang, ``Hand-object contact consistency
  reasoning for human grasps generation,'' in \emph{Proceedings of the IEEE/CVF
  International Conference on Computer Vision}, 2021, pp. 11\,107--11\,116.

\bibitem{MANO:SIGGRAPHASIA:2017}
J.~Romero, D.~Tzionas, and M.~J. Black, ``Embodied hands: Modeling and
  capturing hands and bodies together,'' \emph{ACM Transactions on Graphics,
  (Proc. SIGGRAPH Asia)}, vol.~36, no.~6, Nov. 2017.

\bibitem{hasson19_obman}
Y.~Hasson, G.~Varol, D.~Tzionas, I.~Kalevatykh, M.~J. Black, I.~Laptev, and
  C.~Schmid, ``Learning joint reconstruction of hands and manipulated
  objects,'' in \emph{CVPR}, 2019.

\bibitem{calli2017yale}
B.~Calli, A.~Singh, J.~Bruce, A.~Walsman, K.~Konolige, S.~Srinivasa, P.~Abbeel,
  and A.~M. Dollar, ``Yale-cmu-berkeley dataset for robotic manipulation
  research,'' \emph{The International Journal of Robotics Research}, vol.~36,
  no.~3, pp. 261--268, 2017.

\bibitem{makoviychuk2021isaac}
V.~Makoviychuk, L.~Wawrzyniak, Y.~Guo, M.~Lu, K.~Storey, M.~Macklin,
  D.~Hoeller, N.~Rudin, A.~Allshire, A.~Handa, \emph{et~al.}, ``Isaac gym: High
  performance gpu-based physics simulation for robot learning,'' \emph{arXiv
  preprint arXiv:2108.10470}, 2021.

\end{thebibliography}

\end{document}